# A Hybrid AI Framework for Academic Advising: Integrating Ensemble-Based Grade Prediction and a Rule-Based Expert System

Hamid Saadatfar[1*], Rohollah Hedayati-Nasab[2], AmirHossein Eshghi[3], Arash Hajihashemi[3]

*Email Addresses: {saadatfar@birjand.ac.ir, roh.hedayati@birjand.ac.ir, amir.eshghi@birjand.ac.ir, arashhajihashemi@birjand.ac.ir}*

[1]Associate Professor, Department of Computer Engineering, Faculty of Electrical and Computer Engineering, University of Birjand, Birjand, Iran.

[2]Visiting Lecturer and IT Specialist, Department of Computer Engineering, Faculty of Electrical and Computer Engineering, University of Birjand, Birjand, Iran.

[3]Msc Student, Department of Computer Engineering, Faculty of Electrical and Computer Engineering, University of Birjand, Birjand, Iran.

*: Corresponding author

**Abstract**

The rapidly increasing student population has posed serious challenges to the traditional academic advising process. This study designs and implements a multi-purpose intelligent system to support students' academic progress, based on a two-part hybrid framework: (1) an advanced model for grade prediction and (2) a rule-based recommendation engine. Using a dataset containing 416,558 educational records from the University of Birjand, students were first divided into homogeneous clusters using the Gaussian Mixture Model (GMM). Subsequently, a Stacking Ensemble model combining Random Forest, Gradient Boosting, and MLP was trained specifically for each cluster. Evaluation results demonstrated that the Stacking model outperformed base models across all clusters, achieving a final aggregated RMSE of 2.35. The second component is an expert system that provides intelligent recommendations by synergizing educational regulations with the grades predicted by the first component. This system has been implemented as a practical tool on the University of Birjand portal, offering students real-time feedback such as semester GPA prediction, probation risk warnings, and course suggestions for GPA improvement.

**Keywords:** Educational AI; Academic Advising; Artificial Intelligence; Faculty Students; Expert Systems.

## 1.Introduction

The rapid expansion of higher education systems worldwide has fundamentally transformed the nature of academic advising. Universities are now required to support increasingly diverse student populations with limited human advising resources, while simultaneously ensuring timely guidance, academic progression, and policy compliance. As a result, traditional face-to-face advising models have become difficult to scale, particularly during critical periods such as course registration and probation monitoring.

In recent years, artificial intelligence (AI) and educational data mining techniques have increasingly been adopted to support decision-making in higher education[1-3]. These methods enable the analysis of large-scale academic records to uncover patterns related to student performance, progression, and risk factors. Within academic advising, AI-driven tools have shown promise in assisting advisors and students by predicting outcomes, identifying at-risk learners, and supporting more informed academic planning.

Education as a fundamental human need has always been at the center of attention for societies and policymakers[4]. Improving the quality of education, increasing efficiency, and promoting educational equity are the main concerns in this field; this has led to artificial intelligence entering this field as an extremely powerful tool[5], playing various roles to assist students, professors, advisors, and others. Artificial intelligence has had a significant impact on improving the quality and simplicity of education in various ways and has caused a profound transformation in the field of education[5]; in such a way that the separation between education and artificial intelligence has become impossible, and these two fields are closely intertwined[6]. Academic advising plays a critical role in students' academic success, retention, and timely graduation. However, advising decisions are often based on limited historical information, subjective judgment, or static curricular rules, which may fail to account for heterogeneous learning patterns and evolving student trajectories. This limitation is particularly problematic for students at academic risk, whose early intervention opportunities are frequently missed due to delayed or reactive advising processes.

Recent advances in Artificial Intelligence and Educational Data Mining have introduced powerful tools for predicting academic performance and recommending courses. While machine learning models have demonstrated strong predictive accuracy, most existing systems operate as isolated components—focusing either on prediction, recommendation, or conversational support. More importantly, many AI-driven advising systems function as black boxes, offering limited transparency and weak alignment with institutional regulations and pedagogical principles. This disconnect reduces advisors' trust in automated systems and limits their adoption in real academic environments.

To address these challenges, this study proposes a hybrid academic advising framework that integrates predictive learning analytics with rule-based pedagogical reasoning. Rather than treating grade prediction and advising as separate tasks, the proposed framework unifies data-driven performance modeling with expert-defined academic policies, enabling both accurate prediction and explainable, regulation compliant decision support. By combining clustering-based ensemble learning with a transparent rule-based system, the framework supports proactive, personalized, and scalable academic advising.

The main contributions of this study are summarized as follows:

- A large-scale empirical analysis of academic performance using over 416,000 real student records, enabling robust modeling of heterogeneous learning behaviors.
- A cluster-based ensemble learning framework that captures distinct learner profiles and improves grade prediction accuracy by training specialized models for homogeneous student groups.
- A hybrid advising architecture that integrates machine learning predictions with a rule-based expert system, ensuring transparency, policy compliance, and pedagogical validity.
- A real-world deployment of the proposed system within a university academic portal, demonstrating practical feasibility and supporting proactive academic advising in operational conditions.

The structure of this study is organized as follows. The second section reviews and analyzes the related works in three main areas: advisor systems, course recommendation systems, and student course grade prediction. The third section introduces the data used in the study, describes the preprocessing steps, and explains the proposed method. The fourth section reports the results obtained from the implementation and evaluation of the proposed model of all components of the system. Finally, the fifth section summarizes the findings and presents the main conclusions.

## 2.Related work

In the field of using artificial intelligence in higher education, several applications have attracted the attention of researchers. One of the most important and practical areas among them is academic advising. Academic advising is a very influential sector in students' educational performance. In the past, this process was done traditionally and in person and required a lot of human resources and considerable time. Today with the rapid increase in the number of students, this has become very difficult and impossible. However, with the emergence of artificial intelligence tools and the growing attention toward this field, numerous studies have been conducted on this topic, and many tools have been introduced in this field. One interesting tool among them is chatbots, and in the following, we will examine several examples of them. For example, Mitra Reshmi et al. (2023)[7], developed a chatbot called Ask Rowdy, which was based on IBM Watson Assistant and utilized machine learning algorithms and natural language processing. In this study, data analysis related to emails, phone calls, and in-person meetings was used. The main purpose of this system was to provide real-time responses to frequently asked student questions. The results of this study showed that the use of such chatbots can improve students' access to advisory services and increase the efficiency of the response process. In another study, in 2025 Yixuan Mi et al[8]. designed a system called Smart Course for academic advising of undergraduate Computer Science students. This system is an AI-based, context-aware system that utilizes a large language model (LLM) running locally via Ollama. The main innovation of Smart Course is in integrating three key data sources: student transcripts, the curriculum, and the user's query into a structured, context-aware prompt that is sent to the large language model. The result is relevant, accurate, and actionable recommendations. Among these studies, bilingual examples have also become important. For instance, Sherif Abdelhamid et al. (2025)[9] introduced a web-based platform called Advisely. Advisely integrates a GPT-4 AI-based chatbot with a comprehensive knowledge base that includes the policies, regulations, and requirements of each educational institution. This chatbot was designed using a multi-layered architecture to ensure its modularity, scalability, and ease of maintenance. The system was designed to support students, academic advisors, and faculty members, providing reliable and context-specific textual guidance on topics such as course selection, curriculum requirements, and the interpretation of university policies. Also in 2022, Ghazala Bilquise et al[10]. designed and developed a bilingual chatbot for academic advising that, using artificial intelligence and natural language processing, is capable of answering students' questions in both English and Arabic. The data for this study were collected through direct interaction with students and advisors and official university documents. The chatbot model was constructed using a supervised deep learning algorithm. The main tool for implementing the system was two neural network models (one for English and one for Arabic). The evaluation results showed that the accuracy of the system was about 80% for the English model and about 75% for the Arabic model.

Another problem that students face during their academic career is the correct selection of courses to achieve the best results and greater success in these courses. Today, Artificial Intelligence (AI) is assisting students to make this critical task simpler and more efficient. For example, Atalla et al. [11] presented a three-layer recommendation system to automate academic advising, which is based on analyzing the curriculum structure and students' past performance. This system, using machine learning (ML), graph theory, and Bayesian Belief Networks (BBN), is able to predict a student's future grades and provide personalized course recommendations. The results showed that the proposed system achieved an accuracy of 86% and a mean squared error (MSE = 0.14), performing better than other machine-learning-based methods. In another study, Polyzou et al. [12] presented an academic recommendation system based on session-based recommendation that, beyond conventional methods, not only considers the popularity and sequence of courses, but also pays attention to the relevance and synergy of courses taken in each semester. In this study, two deep learning–based models named CourseBEACON and CourseDREAM were developed, which are capable of modeling intra-semester and inter-semester dependencies by using long short-term memory (LSTM) networks. Also, Wanger et al.[13] presented a K-Nearest Neighbors (KNN)-based course recommendation model aimed at supporting students at risk of dropping out during the early semesters. This model recommends courses that were commonly taken by students who successfully graduated. Data were collected from 1366 students over 6 semesters. The results showed that the system's recommendations had a high overlap with the graduates' educational path, but there was less overlap for dropouts, and this group was usually offered a smaller number of courses (average 3 to 5). Following the recommendations significantly reduced the predicted risk of dropping out and had no negative impact on graduates. Kord et al.[14] in 2025 introduced a multi-model framework consisting of two main components: a grade prediction model and a course recommendation model. This study was conducted using the real-world MU-dataset. The results showed that among the machine learning (ML) and deep learning (DL) algorithms, the Support Vector Classification (SVC) model achieved the best performance in predicting students' academic performance, with an accuracy of 78.04% for multi-class classification (High, Medium, Low) and 75.37% for the F1 score. The recommendation model receives the output of the prediction model (SVC) as input and, based on university regulations, suggests the most appropriate department and courses.

Prediction is one of the most popular and widely used applications of artificial intelligence, implemented through various algorithms. Education is no exception to this trend; in recent years, numerous studies have utilized this capability to predict academic performance and grades. Several of these studies will be discussed in the following. In a study conducted by Mirna Nachouki et al. [15], a model based on the Random Forest algorithm was developed using Educational Data Mining (EDM) techniques to predict students' academic performance and identify the Influencing factors. The data that is used in this research consisted of records from 650 undergraduate students. The results showed that the Random Forest model was able to predict student grades with an accuracy of 90%. Among the limitations of this model are differences in teaching and grading styles of professors, changes in course content, and data imbalances, which can affect the model's accuracy and generalizability. Similarly, Aritra Ghosh et al. [16] introduced the Attentive Knowledge Tracing (AKT) model, which addresses the problem of knowledge tracing to predict learners' future performance based on their past responses. This model combines attention networks, psychometrics, and cognitive science. The results indicated that AKT is both more accurate and more interpretable than traditional models. In addition to higher accuracy in predicting answers, this model is capable of graphically displaying relationships between concepts and analyzing the difficulty level of questions, which is suitable for personalized learning and providing automatic feedback. In another study, Naveed Ur Rehman Junejo et al. [17] proposed a Students Academic Performance Prediction Network (SAPPNet) based on deep learning to predict students' academic performance. This model used combined demographic, psychological, and behavioral data of students before and after the COVID-19 pandemic. The experimental results showed that the proposed model achieved 93% accuracy, recall, and F1-score, outperforming comparative models such as ANN (84%) and SLPNet (89%). In a 2022 study, Yudish Teshal Badal et al.[18] developed a predictive model using the Random Forest algorithm to predict students' academic performance (grades and engagement levels) and analyze the influence of online learning platform features. This framework was implemented as a web-based prototype that users could upload student data files and predict their performance. It is adaptable to various educational institutions and can be deployed accordingly. The results indicated that the system could predict grades with 85.13% accuracy and engagement levels with 83.88% accuracy. Another study employed several machine learning algorithms, including XGBoost (Extreme Gradient Boosting), Logistic Regression, SVM, KNN, and Random Forest, to predict students' GPA (Grade Point Average). In this study, by use early academic data and features. The dataset contained 5,000 student records collected over five consecutive years. Finally, results revealed that the Random Forest algorithm achieved the best performance among all models. The highest accuracy was obtained when admission grades were combined with first-level course grades, while combining admission grades with gender features decreased prediction accuracy across all classifiers. This study was conducted by Essa Alhazmi and Abdullah Sheneamer [19]. Finally, in a 2025 study, Jing Wang and Yun Yu proposed [20] a Logistic Regression Model enhanced with Taylor Expansion. The aim of this study was to present a machine learning approach for predicting students' performance in online learning environments. The data were collected from an online learning platform that contains 300 students enrolled in 15 courses. The proposed model, based on three selected feature indicators, achieved an average accuracy of 0.933 and an F1-score of 0.966 on the test dataset.

Table 1 summarizes the reviewed literature, categorizing them by their primary focus, methodology, and key outcomes. As observed, most existing studies focus on a single aspect either prediction, recommendation, or chatbot interfacing. This research aims to bridge these gaps by integrating a clustering-based ensemble predictor with a rule-based expert system into a unified framework.

**Table1. Summary of Related Works in AI-based Academic Support**

| Category | Author (Year) | Methodology / Algorithm | Key Contribution / Outcome | Limitation / Research Gap |
|---|---|---|---|---|
| Advising Systems | Mitra et al. (2023) [7] | IBM Watson Assistant (NLP) | Automated real-time responses to FAQs via a Chatbot. | Limited to answering static FAQs; lacks personalized grade prediction. |
| | Mi et al. (2025) [8] | LLM (Ollama) + Context Prompting | Integrated transcripts & curriculum for context-aware advice. | Relies heavily on LLM inference which can be computationally expensive; no rule-based safety layer. |
| | Abdelhamid et al. (2025) [9] | GPT-4 + Knowledge Base | Web platform (Advisely) interpreting university policies. | Focuses on policy interpretation rather than performance analytics. |
| | Bilquise et al. (2022) [10] | Deep Learning (RNNs) | Bilingual (English/Arabic) chatbot with ~80% accuracy. | Limited prediction accuracy (75-80%); purely conversational without dashboard analytics. |
| Course Recommendation | Atalla et al. (2023) [11] | Bayesian Belief Networks (BBN) | 3-layer system improving personalization with 86% accuracy. | Bayesian networks can struggle with scalability on very large datasets compared to ensemble methods. |
| | Polyzou et al. (2023) [12] | LSTMs (Deep Learning) | Modeled course dependencies and synergy (CourseBEACON). | Deep learning models often lack explainability ("Black Box" problem) for advisors. |
| | Wagner et al. (2024) [13] | K-Nearest Neighbors (KNN) | Reduced dropout risk by mimicking successful graduates' paths. | Recommendations based solely on "majority success" may not fit unique student profiles. |
| | Kord et al. (2025) [14] | SVC + Rule-Based Engine | Hybrid framework recommending | Accuracy (78%) is lower than ensemble-based methods; limited to |

| | | | depts/courses (78% acc). | classification (High/Med/Low). |
|---|---|---|---|---|
| Grade Prediction | Nachouki et al. (2023) [15] | Random Forest | Identified key predictors with 90% accuracy (small dataset). | Small dataset (650 students) limits generalizability; single-model approach. |
| | Ghosh et al. (2020) [16] | Attention Networks (AKT) | Interpretable knowledge tracing using cognitive science. | Focused on "knowledge tracing" (specific concepts) rather than overall semester GPA/course grades. |
| | Junejo et al. (2024) [17] | Neural Networks (SAPPNet) | 93% accuracy using pre/post-COVID behavioral data. | Requires extensive behavioral/psychological data which is often unavailable in standard SIS. |
| | Badal et al. (2023) [18] | Random Forest | Web tool predicting grades (85%) and engagement (83%). | Does not integrate university regulations (rules) into the prediction pipeline. |
| | Alhazmi et al. (2023) [19] | Random Forest (Comparative) | Confirmed RF superiority on 5,000 records; analyzed demographics. | Did not utilize clustering to handle heterogeneous student data distributions. |
| | Wang & Yu (2025) [20] | Logistic Regression + Taylor Exp. | High precision (F1: 0.96) for online learning performance. | Specific to online learning environments; dataset size (300 students) is very small. |

## 3.Data and methodology

In this section, the data used in the study and the methods employed in the proposed approach are introduced and explained. At first, the collected dataset is described, and details regarding its features, structure, and size are provided. Then, the methodological framework of the research is explained; meaning that the main stages of the research and how to use artificial intelligence models to analyze data and predict results are systematically explained.

### 3.1. Dataset

The data used in this study were collected from the student portal of the University of Birjand. These data include students' academic records in various courses during the period from the first semester of the 2021-2022 academic year to the first semester of the 2024-2025 academic year. Each record represents a single course taken by a student, and the corresponding information recorded in the dataset. The features included in this dataset, along with their corresponding descriptions, are presented in Table 2. In total, 416,558 data records were extracted. This dataset size provides a suitable foundation for using machine learning methods to analyze and predict student performance.

To quantify the linear relationships between variables, the Pearson correlation coefficient was calculated. In the first step, a correlation matrix for all pairs of features was generated (Figure 1) To evaluate the internal dependencies of input variables. In the second step, the correlation coefficient of each feature with the target variable was computed and visualized separately (Figure 2) to provide an initial understanding of how strongly each feature is associated with the model's output.

**Table 2. Description of the Features in the Dataset Used in the Study**

| Feature Name (Standardized) | Value Type | Details |
|---|---|---|
| **Semester Term** | Categorical | Allowed values are 1 and 2. The summer semester has been removed from the data. |
| **Course Credit Hours** | Numerical (Integer) | Total number of units for the course. If the course has practical and theoretical units, their sum is considered. |
| **Course Type Category** | Categorical | Code related to the type of course (e.g., theoretical, practical, workshop, basic). |
| **Instructor History (Mean)** | Numerical (Float) | The average of the selected course with the same professor in past semesters. |
| **Instructor History (Var)** | Numerical (Float) | The variance of the grade for the selected course with the same professor in past semesters. |
| **Student Category GPA** | Numerical (Float) | The student's average grade in courses of the same type in past semesters. |
| **Student Category Var** | Numerical (Float) | The variance of the student's grades in courses of the same type in past semesters. |
| **Gender** | Boolean | Student's gender code (values 1 and 2). |
| **Faculty ID** | Numerical (Integer) | Indicates in which semester of their studies the student is taking this course. |
| **Current Semester Count** | Numerical (Integer) | Total number of units the student has taken up to the semester of taking this course. |
| **Total Credits Attempted** | Numerical (Integer) | Total number of units the student has passed up to the semester of taking this course. |
| **Total Credits Earned** | Numerical (Float) | Student's previous semester GPA. For the first semester, the GPA from the second semester of the prior academic year is used. Summer GPAs are excluded. |
| **Previous Term GPA** | Numerical (Float) | The student's cumulative GPA up to the semester of taking this course. |
| **Cumulative GPA (CGPA)** | Numerical (Integer) | Total number of units the student is taking in the current semester. |
| **Current Semester Load** | Categorical | The code for the course taken by the student. |
| **Course ID** | Categorical | The code for the professor teaching the course. |
| **Instructor ID** | Categorical | The student's field of study code. |
| **Major Field ID** | Numerical (Float) | The grade obtained by the student in the course (this is the target variable). |
| **Final Grade** | Numerical (Float) | The grade obtained by the student (Target Variable). |

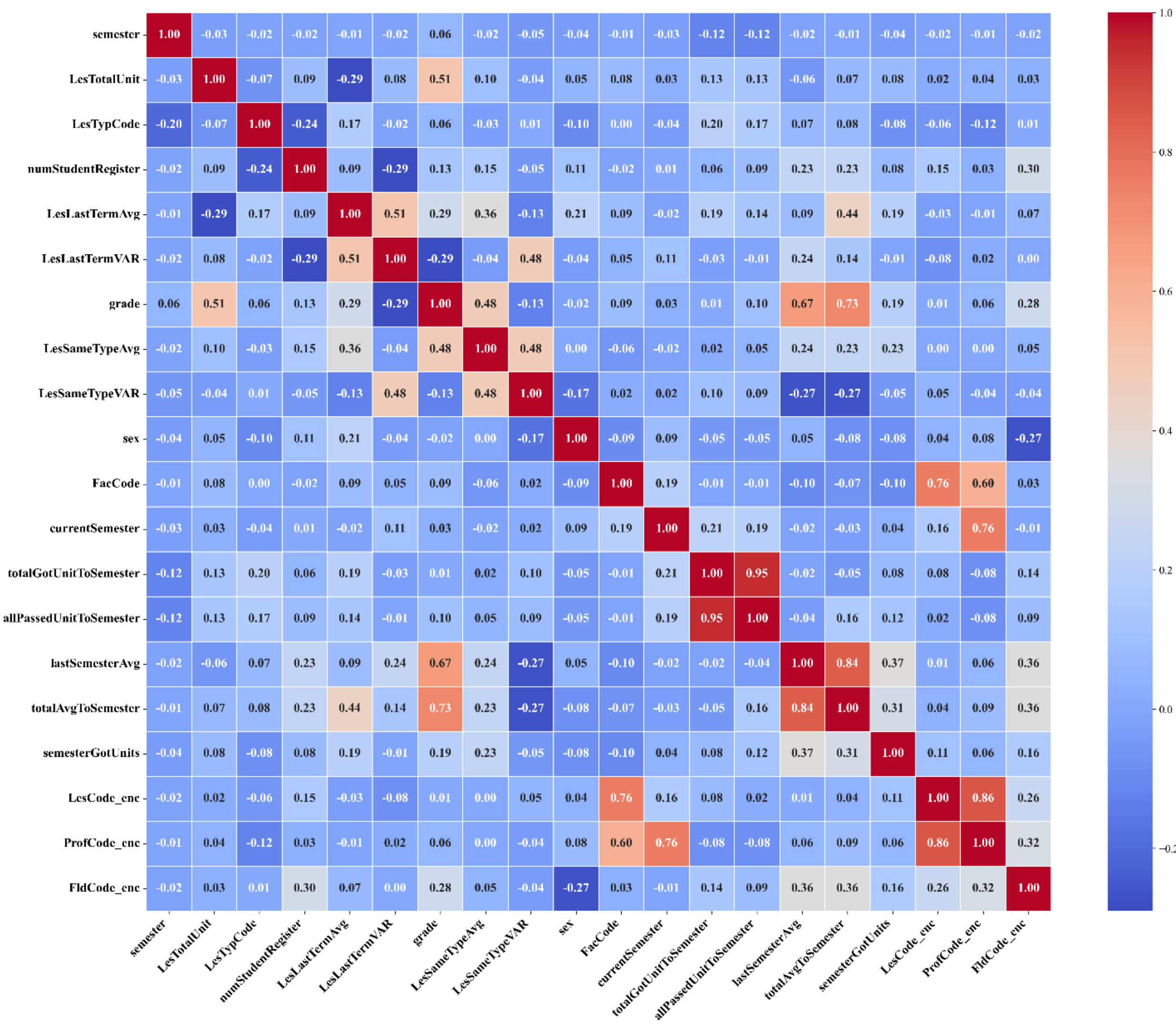

**Figure 1. Pearson correlation matrix between all numerical features in the dataset.**

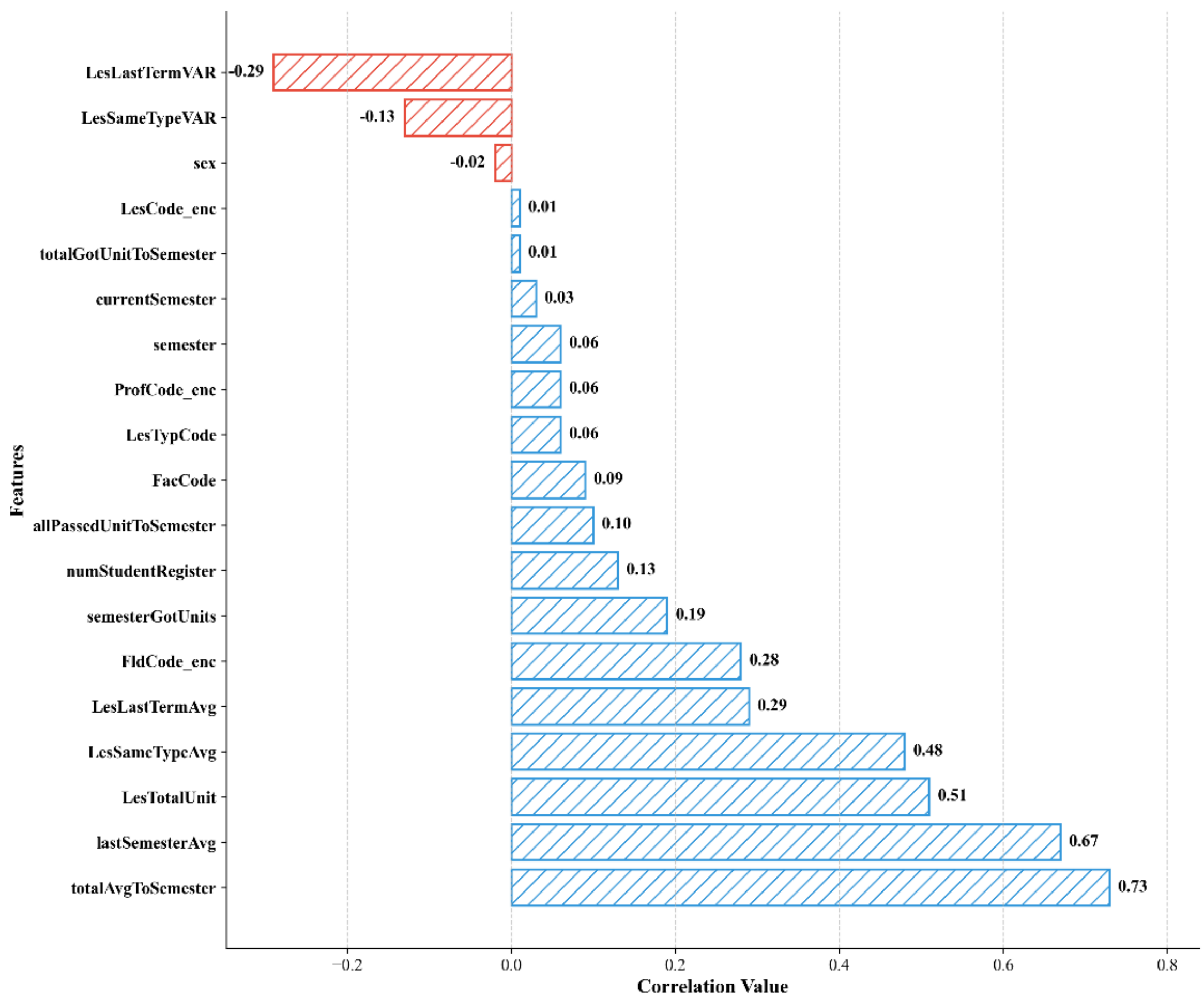

**Figure 2. Pearson correlation of each feature with the target variable (course grade).**

### 3.2. Framework

This study introduces an intelligent multi-purpose system that supports and guides students in their academic journey. The system is based on two main components: (1) an advanced hybrid framework for predicting academic grades, and (2) a rule-based recommendation engine for providing educational suggestions.

The first part of this research focuses on using a hybrid model to predict course grades. The main idea of this framework is to divide the complex data space into more homogeneous subsets through clustering and then train three regression models for each subset. This approach aims to enhance prediction accuracy. Figure 1 shows the structure of proposed method for prediction, which consists of two main stages: (1) data clustering and (2) training ensemble regression models for each cluster.

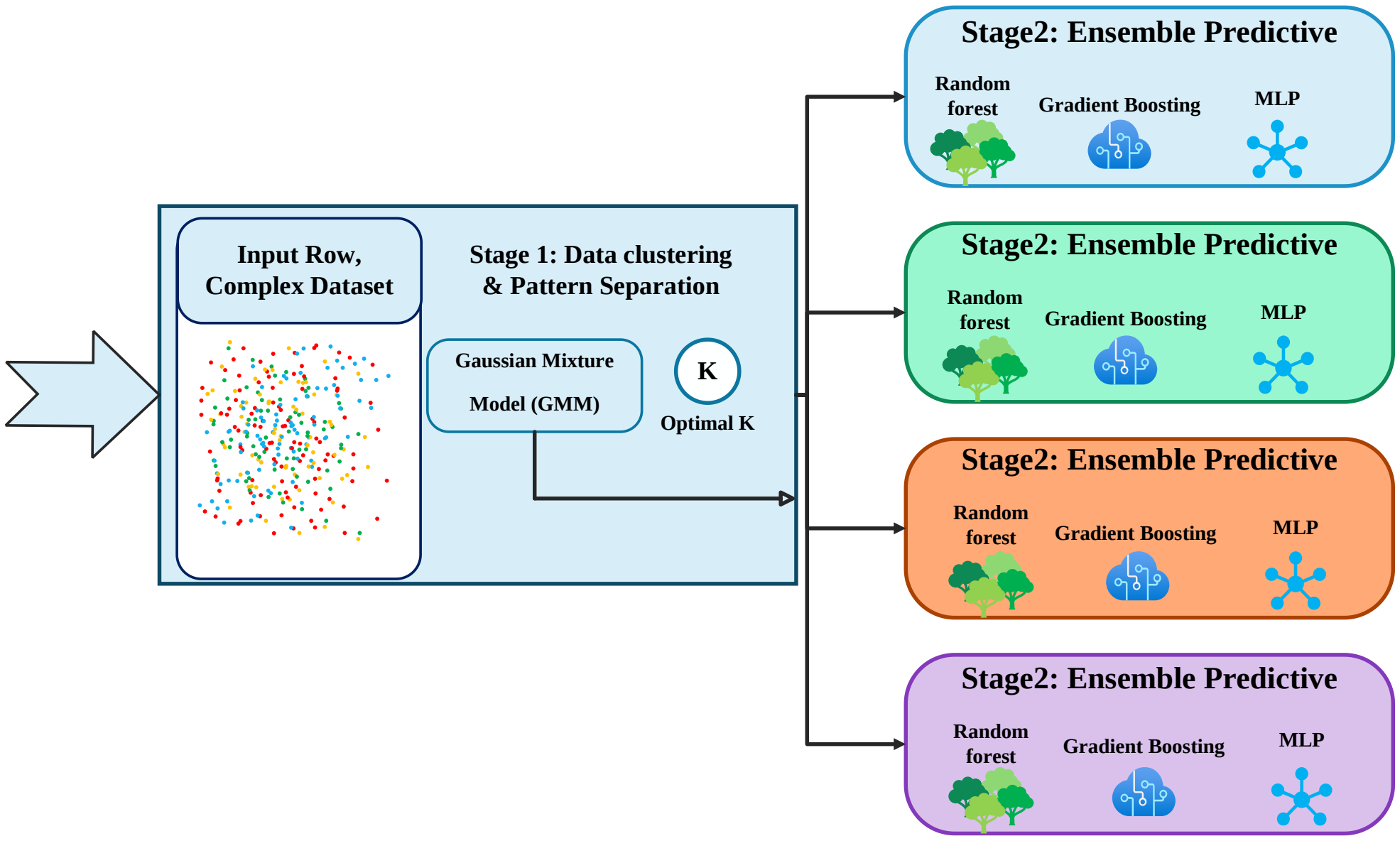

**Figure 3. General framework for predicting course grades**

**Stage One: Data clustering for pattern segmentation:**

In the first stage, the goal is to identify and distinguish similar statistical patterns among students. For this purpose, the Gaussian Mixture Model (GMM) algorithm is employed. GMM is selected due to its ability to model clusters with complex probabilistic structures. To ensure the optimal cluster structure, the optimal number of clusters (K) is determined through an evaluation process based on K-Fold cross-validation.

**Stage Two: Development of ensemble prediction models**

After dividing the data into K homogeneous clusters, in this stage, an independent ensemble regression model is trained for each cluster. This ensemble model is constructed by combining the outputs of three powerful base learners:

- Random Forest: This method is based on aggregating multiple decision trees, where each tree is trained on a random subset of the data and features. This mechanism reduces variance and prevents overfitting, resulting in more stable and accurate predictions[21].
- Gradient Boosting: It is an incremental reinforcement algorithm that trains weak models (typically shallow decision trees) in a stage-wise manner. At each stage, the remaining errors from the previous stage are modeled to gradually improve overall performance. This property enables Gradient Boosting to effectively learn complex patterns[22].
- Multi-Layer Perceptron (MLP): a feedforward neural network consisting of multiple hidden layers with nonlinear neurons. This structure can identify and model complex nonlinear relationships between variables and it is especially efficient in problems with nonlinear data[23].

To achieve the final prediction of each cluster, the outputs of the three base models are combined using the averaging method. This strategy uses the structural and behavioral diversity of the models to help reduce overall error, increase stability, and significantly improve the final accuracy of the framework.

The second part of the study includes the development of a Rule-Based System. This system alone can provide students with recommendations for course registration in upcoming semesters based on predefined rules. In the latest capability, the system is able to generate smarter suggestions by combining academic rules (part two) and predicted scores (part one). In other words, the system recommends courses that the student is not only eligible to take but also has a high probability of achieving the desired grade in them, based on the prediction model. This synergy helps students pursue a more optimal and successful academic path.

### 3.3. Hyperparameter Optimization

To ensure optimal performance and reproducibility, the hyperparameters for all base models (Random Forest, Gradient Boosting, and MLP) were optimized using a Grid Search approach coupled with 5-Fold Cross-Validation. This rigorous process involved defining a search space for key parameters and selecting the combination that minimized the prediction error (RMSE) on the validation set. The detailed configuration of the search space and the optimal parameters selected for each model are presented in Table 3.

**Table 3. Hyperparameter search space and optimal values selected via Grid Search**

| Model | Hyperparameter | Search Space (Grid) | Optimal Value |
|---|---|---|---|
| **Random Forest** | n_estimators | [50, 100, 200, 300] | 100 |
| | max_depth | [10, 20, 30, None] | None (Full) |
| | min_samples_split | [2, 5, 10] | 2 |
| **Gradient Boosting** | n_estimators | [50, 100, 200] | 100 |
| | learning_rate | [0.01, 0.05, 0.1, 0.2] | 0.1 |
| | max_depth | [3, 5, 7] | 3 |
| **MLP Regressor** | hidden_layer_sizes | [(64,32), (128,64), (256,128,64)] | (256, 128, 64) |
| | activation | ['relu', 'tanh', 'logistic'] | relu' |
| | max_iter | [200, 300, 500] | 300 |

## 4.Results and Implementation

In this section, the outputs and findings of the proposed framework are presented and analyzed. The aim is to demonstrate the model's effectiveness across different stages, including data preprocessing, the predictive performance of the hybrid model, the examination of a case study, and finally, the development of a practical toolbox for the grade prediction and a recommendation system based on predicted grades and students' performance. This structure allows for a step-by-step analysis of the results and provides a clearer explanation of the role and impact of each component within the proposed approach.

### 4.1. Preprocessing

In this study, data preprocessing was carried out in a step-by-step and systematic manner to ensure data quality and the compatibility of samples with machine learning processes. Suppose we represent the raw dataset by $D = \{(x_i, y_i)\}_{i=1}^{N}$ where $x_i$ is the feature vector of sample $i$ and $y_i$ is the target value. The preprocessing steps are as follows.

I. Removing zero and null values

Some specific columns related to domain-specific assumptions (e.g., Previous Semester GPA, Course GPA in prior semesters, and GPA of similar courses), contained zero or null values, which in the context of our data indicate the absence of a value (for example, first-semester students or courses being offered for the first time), not an actual score. To prevent noise or bias from entering the learning process, all rows that contained at least one zero or null value in any of these key columns were removed.

II. Encoding of Categorical Variables

Three categorical columns —course code, instructor code, and major code—were encoded for use in machine learning models, and the resulting mappings were stored to ensure consistency during evaluation and deployment. The following two methods were used for coding (selected based on the number of unique values and the complexity of each column):

- **One-Hot Encoding** – For columns with a relatively small number of categories (such as major code), each category c is converted into a binary vector $\phi OHE(c_j) \in \{0,1\}^K$ ,where K is the total number of categories, and only one entry corresponding to that category is equal to 1.
- **Target / Mean Encoding**, or smoothing with cross-validation, was used for columns with a large number of categories (such as instructor ID) to prevent the feature space from becoming too high-dimensional. For each category c the mean value of the target in that category is computed and then smoothed with the global mean to reduce noise caused by categories with few samples. The general form of the smoothing is as follows (Formula 1):

$$TE(c) = \frac{n_c.\mu_c + k.\mu_{global}}{n_c + k} \quad (1)$$

Where $n_c$ is the number of samples in category c and $\mu_c$ is the mean target value in category c, $\mu_{global}$ is the global mean target value, and k is the smoothing parameter.

III. Normalization

To normalize the numerical features (such as GPAs and other quantitative values), z-score standardization was applied [24] so that each feature attains a mean of zero and a standard deviation of one (Formula 2).

$$z_i = \frac{x_i - \mu}{\sigma} \quad (2)$$

In the above formula, $\mu$ represents the mean of each feature and $\sigma$ represents the standard deviation of each feature. In practice, the StandardScaler implementation from sklearn.preprocessing in Python was used, which applies exactly the same formula. The computed values on the training set were stored, and the same parameters were applied to the validation and test data to prevent information leakage.

### 4.2. Creating a hybrid model

After the preprocessing stage, the dataset was subjected to a clustering process to identify similar statistical patterns among the samples. For this purpose, the Gaussian Mixture Model (GMM) was employed [25] as the clustering method, cause this model is a suitable choice due to its flexibility in modeling data distributions as a weighted combination of multivariate Gaussian components.

To determine the optimal number of clusters, the model was fitted over a range of cluster counts. Then, the performance of each model was evaluated using two widely used model selection metric for mixture models, namely the Bayesian Information Criterion (BIC) (Equation 1) and the Akaike Information Criterion (AIC) (Equation 2).

$$BIC = k.\ln(n) - 2l(\hat{\theta}) \quad (3)$$

$$AIC = 2k - 2l(\hat{\theta}) \quad (4)$$

Both criteria establish a balance between model fit and simplicity by penalizing overly complex models. The model with the lowest BIC/AIC value was selected as the optimal solution. Based on this evaluation, the optimal number of clusters in the dataset was determined to be ten. After that, the dataset was divided into training (80%) and testing (20%) subsets, And the clustering process was applied consistently to both subsets to keep the data structure. In addition, K-Fold cross-validation was used into the pipeline to ensure the stability of the clustering results and reduce sensitivity to random initialization.

As explained in Section Three, three base regressors were trained for each identified cluster. These models include the Random Forest Regressor, Gradient Boosting Regressor, and Multi-Layer Perceptron Regressor. The details of each model are presented as follows.

i. Random Forest Regressor:
   This model is an ensemble learning method based on decision trees. In this project, 100 decision trees were used (n_estimators = 100). Random Forest reduces overfitting by constructing multiple independent decision trees and combining their predictions through averaging. The use of random feature sampling at each node increases diversity between the trees and causes more stable predictions.
ii. Gradient Boosting Regressor:
   Gradient Boosting is an additive boosting method in which models are built sequentially. In this project, 100 weak decision-tree learners were used. In each iteration, a new tree predicts the residual error of the previous model, and in this way, the final model is formed through the gradual reduction of error. The distinguishing feature of this method is its high accuracy and ability to model complex and nonlinear relationships; however, if the parameters are not properly tuned, there is a risk of overfitting.
iii. Multi-Layer Perceptron Regressor (MLP):
   The multilayer perceptron network is a deep learning model based on artificial neural networks. In this project, a three-hidden-layer architecture with 256, 128, and 64 neurons was used, and the maximum number of training iterations was set to 300. this network is able to learn complex patterns and nonlinear relationships between variables by employing nonlinear activation functions. An important feature of MLP is its high flexibility in approximating complex functions; however, its training requires more time and is sensitive to the scale of the data, which was handled through data normalization.

After training the base models, their results were combined using the Averaging Stacking Ensemble method. In this approach, the outputs of individual models are averaged and finally produce the prediction. The main advantage of the averaging stacking technique is that it covers the weaknesses of individual models, and increasing the overall accuracy and stability of predictions. Finally, one averaging stacking model was constructed for each cluster; thus, ten final ensemble models were obtained for the entire dataset.

1. Evaluation Methods
   In this study, several evaluation metrics were used to compute the accuracy and error of the regression models. One of the most important metrics is the Mean Squared Error (MSE) [26], and its formula is presented in Equation (5). This metric is more sensitive to large errors due to the squaring of errors, making it highly suitable for models where managing large errors is important. Also, MSE is widely recognized as one of the most commonly used metrics for evaluating the performance of regression models.

$$MSE = \frac{1}{N}\sum_{i=1}^{N}(y_i - \hat{y}_i)^2 \tag{5}$$

Another metric used in this study is the Root Mean Squared Error (RMSE)[27], and its formula is presented in Equation (6). Unlike MSE, which is expressed in the squared unit of the target variable, RMSE is reported directly in the same unit as the target. This property makes the interpretation and understanding of this metric easier for evaluating model performance.

$$RMSE = \sqrt{MSE} \tag{6}$$

Finally, another useful metric for evaluating model performance is the R-Score [27], or coefficient of determination, and its formula is presented in Equation (7). This metric, recognized as a standard for model evaluation, measures the performance of the model based on the proportion of variance in the data explained and is very useful in analyzing the results.

$$R^2 = 1 - \frac{\sum_{i=1}^{N}(y_i - \hat{y}_i)^2}{\sum_{i=1}^{N}(y_i - \bar{y}_i)^2} \tag{7}$$

In formulas 5 and 7, N represents the total number of available data points, $y$ is the true value of the target variable, $\hat{y}$ is the value predicted by the trained model, and $\bar{y}$ represents the mean of the target variable used in the calculations.

## 4.Results

Initially, the results of each base model for each of the extracted clusters were computed, and the detailed outcomes are presented separately in Table 4.

**Table 4. The results obtained from each base model on each cluster, reported separately**

| Cluster | Model | MSE | RMSE | R2 |
|---|---|---|---|---|
| 0 | RandomForest | 4.62 | 2.15 | 0.41 |
| 0 | GradientBoosting | 5.09 | 2.26 | 0.25 |
| 0 | MLP | 6.58 | 2.58 | 0.15 |
| 1 | RandomForest | 4.25 | 2.06 | 0.44 |
| 1 | GradientBoosting | 4.66 | 2.16 | 0.29 |
| 1 | MLP | 5.84 | 2.42 | 0.22 |
| 2 | RandomForest | 6.95 | 2.64 | 0.29 |
| 2 | GradientBoosting | 7.17 | 2.68 | 0.26 |
| 2 | MLP | 10.77 | 3.28 | -0.11 |
| 3 | RandomForest | 6.22 | 2.52 | 0.42 |
| 3 | GradientBoosting | 6.57 | 2.57 | 0.24 |
| 3 | MLP | 10.21 | 3.22 | 0.09 |
| 4 | RandomForest | 5.83 | 2.42 | 0.44 |
| 4 | GradientBoosting | 8.29 | 2.51 | 0.4 |
| 4 | MLP | 8.47 | 2.91 | 0.19 |
| 5 | RandomForest | 4.95 | 2.22 | 0.47 |
| 5 | GradientBoosting | 5.61 | 2.37 | 0.4 |
| 5 | MLP | 7.2 | 2.68 | 0.22 |
| 6 | RandomForest | 5.2 | 2.43 | 0.47 |
| 6 | GradientBoosting | 6.4 | 2.53 | 0.42 |
| 6 | MLP | 9.9 | 3.15 | 0.11 |
| 7 | RandomForest | 6.57 | 2.56 | 0.48 |
| 7 | GradientBoosting | 7.31 | 2.68 | 0.44 |
| 7 | MLP | 10.77 | 3.28 | 0.16 |
| 8 | RandomForest | 6.15 | 2.48 | 0.57 |
| 8 | GradientBoosting | 6.62 | 2.57 | 0.53 |
| 8 | MLP | 9.73 | 3.12 | 0.31 |
| 9 | RandomForest | 5.77 | 2.46 | 0.46 |
| 9 | GradientBoosting | 6.09 | 2.47 | 0.41 |
| 9 | MLP | 9.21 | 3.05 | 0.09 |

To provide a more precise comparison of the performance of the models in each cluster with the constructed ensemble model, reports were generated using various metrics. The ensemble model was developed by combining and averaging the outputs of the base models. In Figure 2, the error values based on the MSE metric are presented for each cluster. As observed, the ensemble model achieved the lowest MSE compared to the other models. In Figure 3, for better interpretation of the results, error values based on the RMSE metric are shown, which provides a more realistic representation of each model's error, as it considers the units of the data. Finally, Figure 6 reports the $R^2$ value, which represent the accuracy and explanatory power of each model in predicting the data. According to these results, the ensemble model outperforms all other models across all metrics.

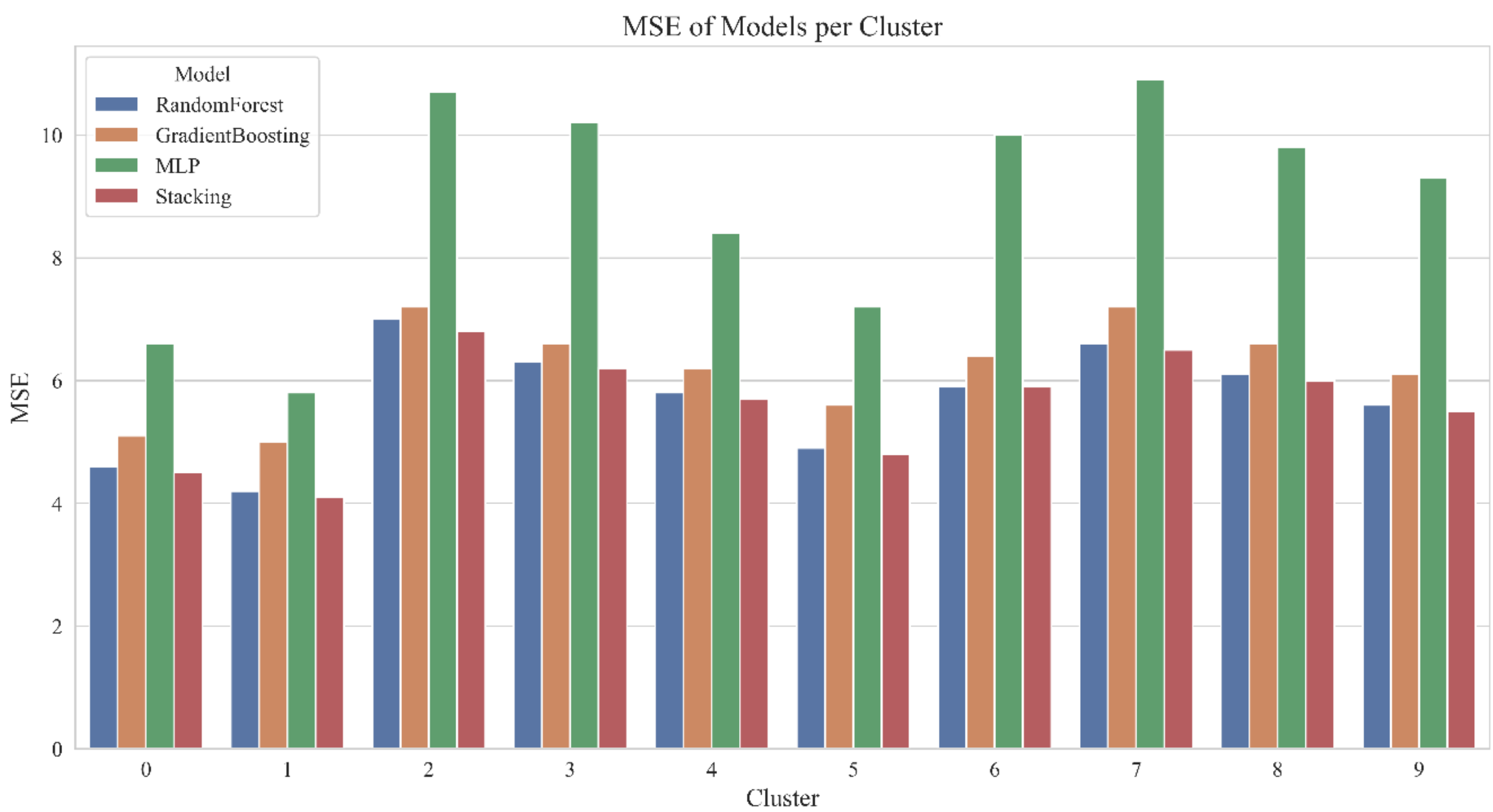

**figure 4. Comparative chart of MSE errors for the base models (RF, GB, MLP) and the ensemble model (AVG Stacking) across all clusters**

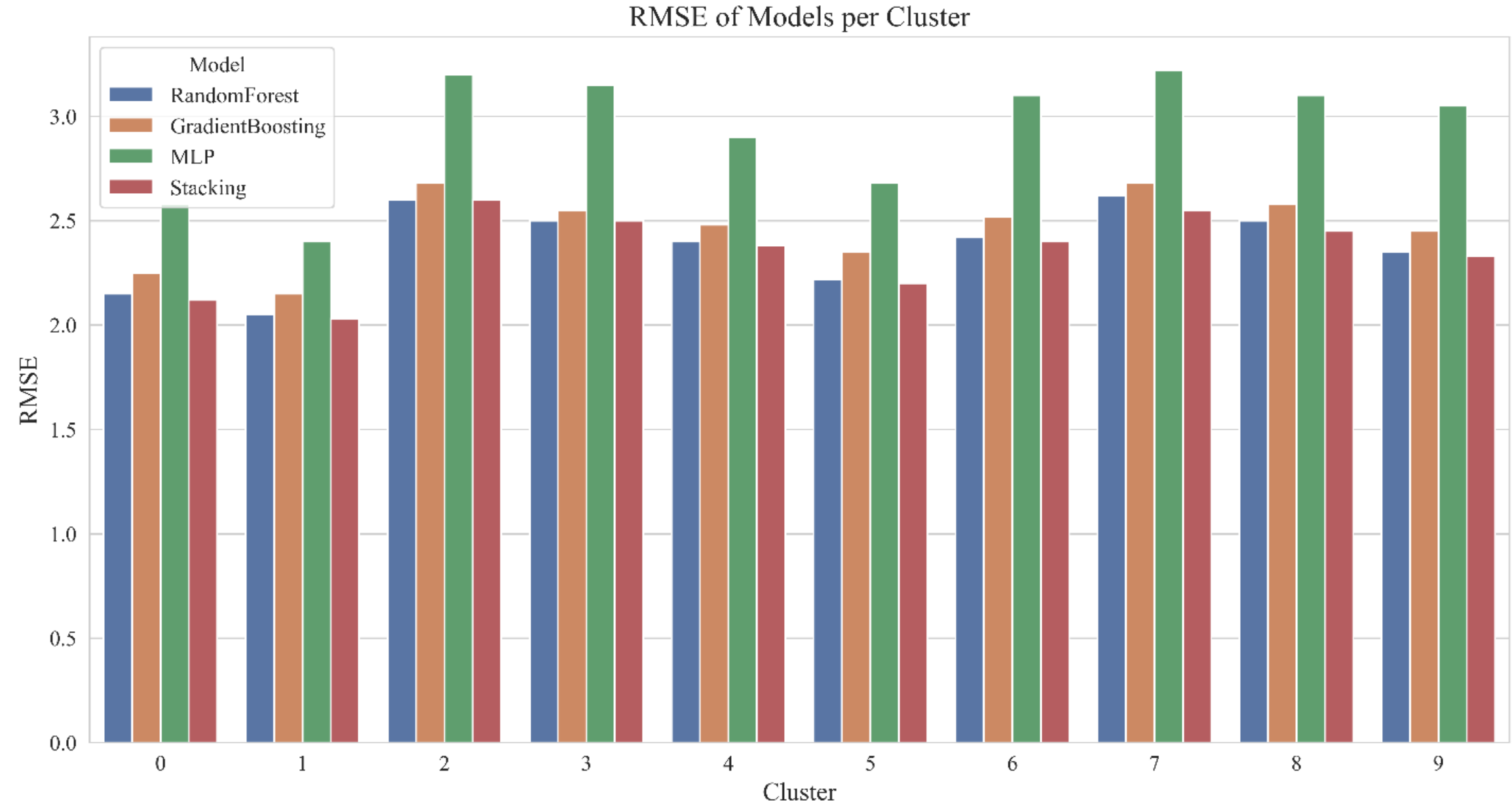

**figure 5. Comparative chart of Root Mean Square Error (RMSE) for the base models and the ensemble model (AVG Stacking) across all clusters**

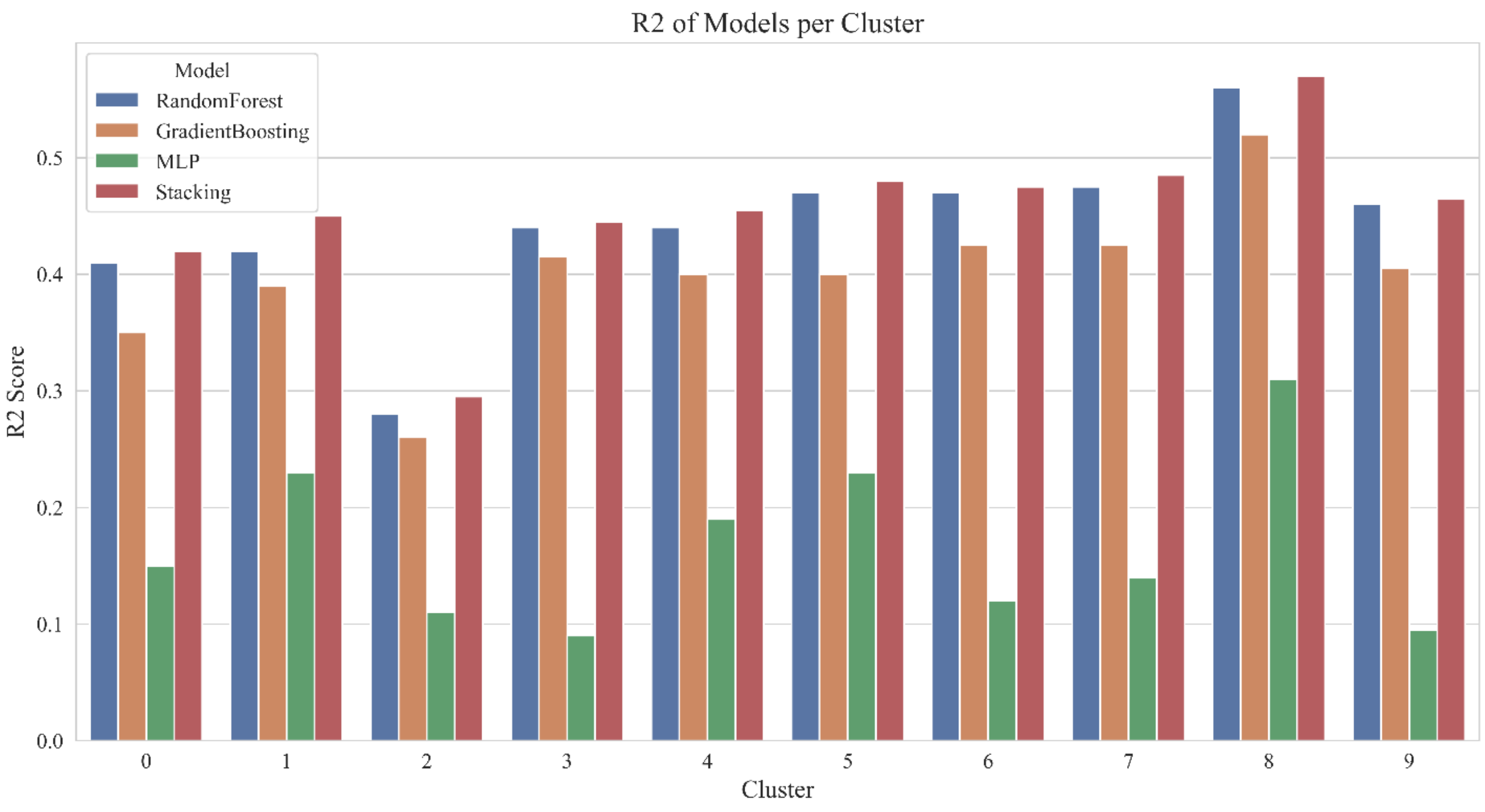

**Figure 6. Comparative $R^2$ chart for the base models and the stacked (ensemble) model across all clusters**

Finally, the performance of the proposed ensemble model for each identified cluster is detailed in Table 5. This table reports the accuracy and error metrics (MSE, RMSE, and R2) at the cluster level, demonstrating how specialized models adapt to specific student sub-groups. Furthermore, to validate the necessity of the clustering phase, the overall performance of the proposed framework was aggregated and compared against a baseline Global Stacking model (trained without clustering). These comparative results are presented in Table 6.

**table 5. Final performance results of the proposed model (Stacking) for each of the 10 clusters**

| Cluster | Model | MSE | RMSE | R2 |
|---|---|---|---|---|
| **0** | Avg Stacking | 4.57 | 2.14 | 0.43 |
| **1** | Avg Stacking | 4.19 | 2.05 | 0.45 |
| **2** | Avg Stacking | 6.84 | 2.61 | 0.3 |
| **3** | Avg Stacking | 6.26 | 2.5 | 0.45 |
| **4** | Avg Stacking | 5.7 | 2.39 | 0.46 |
| **5** | Avg Stacking | 4.86 | 2.21 | 0.48 |
| **6** | Avg Stacking | 5.83 | 2.41 | 0.48 |
| **7** | Avg Stacking | 6.48 | 2.54 | 0.49 |
| **8** | Avg Stacking | 6.05 | 2.46 | 0.57 |
| **9** | Avg Stacking | 5.51 | 2.35 | 0.46 |

**Table 6. Comparison of the aggregated results of the proposed Cluster-Based Stacking model versus the Global Stacking model (without clustering) on the test dataset.**

| Model | MSE | RMSE | R2 |
|---|---|---|---|
| Proposed Method (Cluster-Based Stacking) | 5.53 | 2.35 | 0.53 |
| Global Stacking (Without Clustering) | 6.13 | 2.48 | 0.47 |

As evident from the results (Table6), the clustering process has made a significant contribution to improving the accuracy of the predictions compared to the non-clustered method.

### 4.1. Statistical Significance Analysis

To verify that the performance improvement achieved by the proposed Stacking Ensemble framework is statistically significant and not merely a result of random chance, a Paired Samples T-Test was conducted on the unseen test dataset. Prior to the analysis, the normality of the prediction errors (residuals) was verified using the Shapiro-Wilk test, ensuring that the assumptions of the T-test were met. The null hypothesis ($H_0$) posits that there is no significant difference between the mean errors of the base model and the proposed ensemble model, while the alternative hypothesis ($H_1$) asserts that the proposed framework yields a statistically significant reduction in error. asserts that the proposed framework yields a statistically significant reduction in error. The test was performed at a significant level of $\alpha = 0.05$. The results yielded a p-value < 0.001. Since the p-value is significantly lower than the threshold of 0.05, the null hypothesis is rejected. This confirms that the reduction in RMSE achieved by the Stacking Ensemble represents a statistically significant improvement in predictive accuracy.

### 4.2. Sample Implementation

In this section, to evaluate the model's performance, a number of student samples were entered into the system for testing. For this purpose, three students were randomly selected from the dataset and used as inputs to the model. Figure 7 shows the model's output for these three student samples.

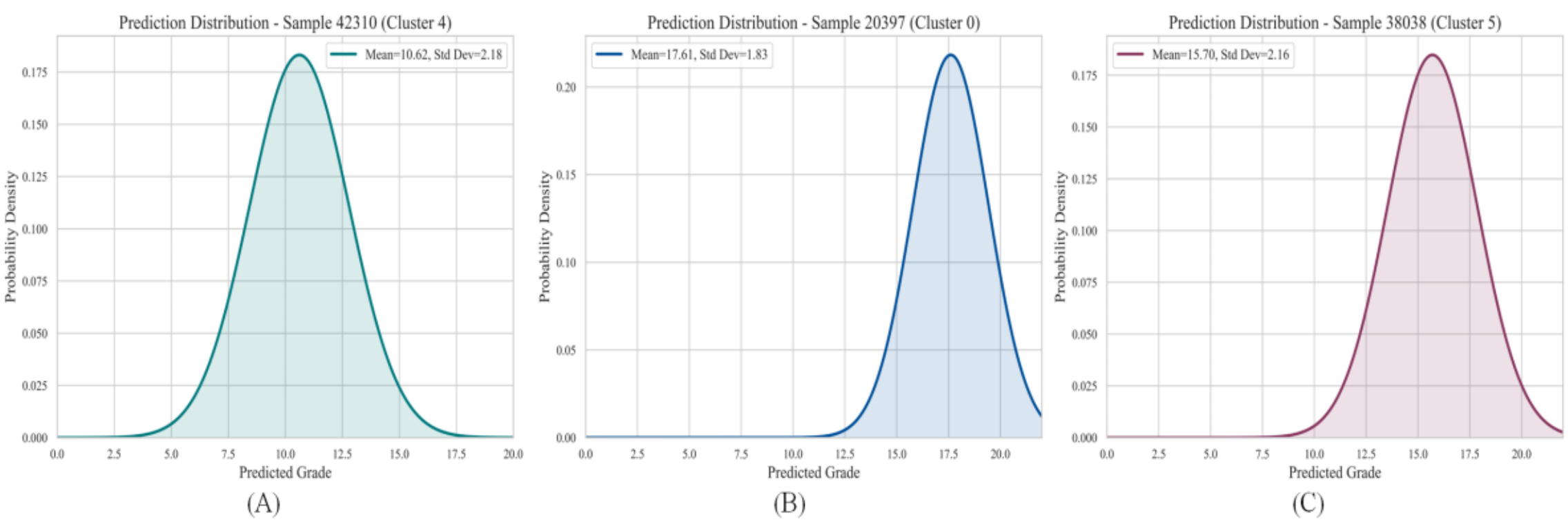

**Figure 7. Displaying the predicted grade probability distribution for three randomly selected student samples from different clusters (A, B, and C).**

Figure 7 shows the predicted grade probability distribution for three selected students from different clusters. In sample (A), belonging to cluster 4, the mean grade is 10.62 with a standard deviation of 2.18, indicating a higher likelihood of obtaining a mid-range grade with relative uncertainty. In sample (B) from cluster 0, the mean grade is 17.61 with a standard deviation of 1.83, showing a more concentrated distribution around higher grades, which reflects the model's higher confidence in predicting good student performance. Finally, sample (C) from cluster 5 has a mean of 15.10 with a standard deviation of 2.16, and its symmetric distribution shows the highest probability around grade 15, with more dispersion compared to cluster 0. Overall, the comparison of these three examples shows that the model is able to provide distinct predictions in terms of mean score and level of uncertainty, depending on the characteristics of each cluster.

### 4.3. Rule-Based Recommendation System

In this section, one of the main features of the system, namely the intelligent recommendation system, is described. This system is designed to provide students with instant, personalized, and proactive feedback during course registration and throughout their academic journey. Unlike conventional recommender systems that operate solely on machine learning, this project uses a combination of a rule-based (Expert System) approach with machine learning. This approach was chosen due to its high transparency, full explainability, and the ability to directly implement university regulations and academic rules. This system generates a set of warnings, recommendations, and motivational messages by analyzing the student's current and past academic status and considering their ongoing course selection. In the following, the architecture, logic, and rules implemented in this system are explained in detail.

The process works as follows: when a student enters the course registration system or after finalizing their course selections, their data is extracted from the fact database and matched against all the rules in the knowledge base by the inference engine. The output of this process is a list of messages appropriate to the student's status that is displayed to them.

#### 4.3.1. Detailed explanation of the implemented rules

In the following, each of the rules defined in the system's knowledge base, along with the logic and objective of each, is described.

- Group 1: Regulation-Based Rules

  These are strict, non-negotiable university regulations that the system validates.

- Group 2: Proactive Warning Rules

These are intelligent rules that warn the student based on data-driven analysis (such as downward performance trends, risk of probation, or slow academic progress).

- Group 3: Smart Recommendation Rules

  This section the intelligent core of the system is responsible for generating personalized recommendations.

- Group 4: Motivational and Feedback Rules

  This section covers the psychological aspect and the maintenance of student motivation.

Table 7 provides a complete description of the rules.

**Table7. A detailed description of the rules, conditions, and actions defined in the knowledge base of the recommendation system.**

| Category | Rule Description | Trigger Condition (IF) | System Action (THEN) |
|---|---|---|---|
| Policy Compliance | Academic Probation Credit Cap | LastTermGPA < 12 (On Probation) AND CurrentSemesterCredits > 14 | "As you are on academic probation, you cannot take more than 14 credits. Please amend your registration." |
| Policy Compliance | Prerequisite & Co-requisite Validation | A selected course's prerequisites are not met OR its co-requisites are not taken concurrently. | "Prerequisite for 'Course_Name' is not met. Please review the curriculum." |
| Proactive Warning | Declining GPA Trend Detection | LastTermGPA < PreviousTermGPA | “A declining GPA trend has been detected. Please consult your academic advisor." |
| Proactive Warning | Predicted Probation Risk | PredictedCurrentTermGPA < 12 | "Your predicted GPA for this term is below 12.0. We strongly recommend reducing your course load and consulting your advisor." |
| Proactive Warning | Slow Academic Progression | AveragePassedCreditsPerTerm < 14 | "Your low average of passed credits may delay your graduation. If not on probation, consider taking at least 17 credits." |
| Intelligent Advisory | Smart GPA Enhancement | (A) PredictedCurrentTermGPA < 14 (At-risk)<br><br>(B) All other students | (A) Suggest 5 eligible courses with PredictedGrade > 14.<br><br>(B) Suggest 3 eligible courses where PredictedGrade > PredictedCurrentTermGPA. |
| Motivational | Positive Reinforcement | LastTermGPA > PreviousTermGPA | "Congratulations! Your academic performance shows a positive upward trend. Keep up the excellent work!" |

**Algorithm 1 presents the pseudocode of the recommendation system.**

```
Algorithm 1
FUNCTION GenerateRecommendations(studentProfile):
    // Initialize an empty list to store messages
    messages = []

    // Rule 1: Check for academic probation credit limit
    IF studentProfile.lastTermGPA < 12 AND studentProfile.currentCredits > 14 THEN
        ADD "Warning: As you are on academic probation, you cannot take more
            than 14 credits." TO messages
    END IF


    // Rule 2: Verify course prerequisites
    FOR EACH course IN studentProfile.currentCourses:
        IF hasUnmetPrerequisites(course, studentProfile.passedCourses) THEN
            ADD "Warning: Prerequisite for the course '" + course.name +
                "' is not met." TO messages
        END IF
    END FOR

    // Rule 3: Detect declining GPA trend
    IF studentProfile.gpa_Term_T_minus_1 < studentProfile.gpa_Term_T_minus_2 THEN
        ADD "Recommendation: Your GPA has been declining. Please consult your
            academic advisor." TO messages
    END IF

    // Rule 4: Warn about critical predicted GPA
    IF studentProfile.predictedGPA < 12 THEN
        ADD "Alert: Your predicted GPA is below 12.0. We strongly recommend
            visiting your advisor..." TO messages
    END IF

    // Rule 5: Recommend courses to improve GPA
    IF studentProfile.predictedGPA < 14 THEN
        suggestedCourses = findCourses(studentProfile,
        criteria='high_predicted_score_above_14', limit=5)

        ADD "Suggestion: To increase your GPA, consider...: " + suggestedCourses TO
            messages
    ELSE
        suggestedCourses = findCourses(studentProfile, criteria =
                'score_higher_than_predicted_gpa', limit=3)
        IF suggestedCourses IS NOT EMPTY THEN
            ADD "Suggestion for GPA enhancement: Consider...: "
                + suggestedCourses TO messages
        END IF
    END IF


    // Rule 6: Check for slow academic progress
    averageCreditsPerTerm = studentProfile.totalPassedCredits /
                studentProfile.numberOfTerms
    IF averageCreditsPerTerm < 14 THEN
        ADD "Warning: Your low rate of passed credits might delay your
```

```
                    graduation..." TO messages
      END IF


      // Rule 7: Provide positive reinforcement for improving GPA
      IF studentProfile.gpa_Term_T_minus_1 > studentProfile.gpa_Term_T_minus_2
      THEN
                    ADD "Congratulations! Your academic performance has improved...
                         Keep up the excellent work!" TO messages
      END IF
      RETURN messages
```

### 4.4. The developed tool implemented on the Birjand University integrated system

In the PUYA system of Birjand University, a new feature has been developed within the student course registration section. During the course registration period, students who have registered for at least 12 credits can use the intelligent assistant. By clicking the corresponding option, the student's selected course information is sent to an API implemented in Python. Then, The system uses the proposed models to calculate the predicted grade for each selected course and presents the results to the user in a structured report. Figure 8 shows a sample from the university system, where the predicted grades for two selected courses of a student from the dataset are displayed. Additionally, the system calculates and provides the student's probable current semester GPA based on these predictions.

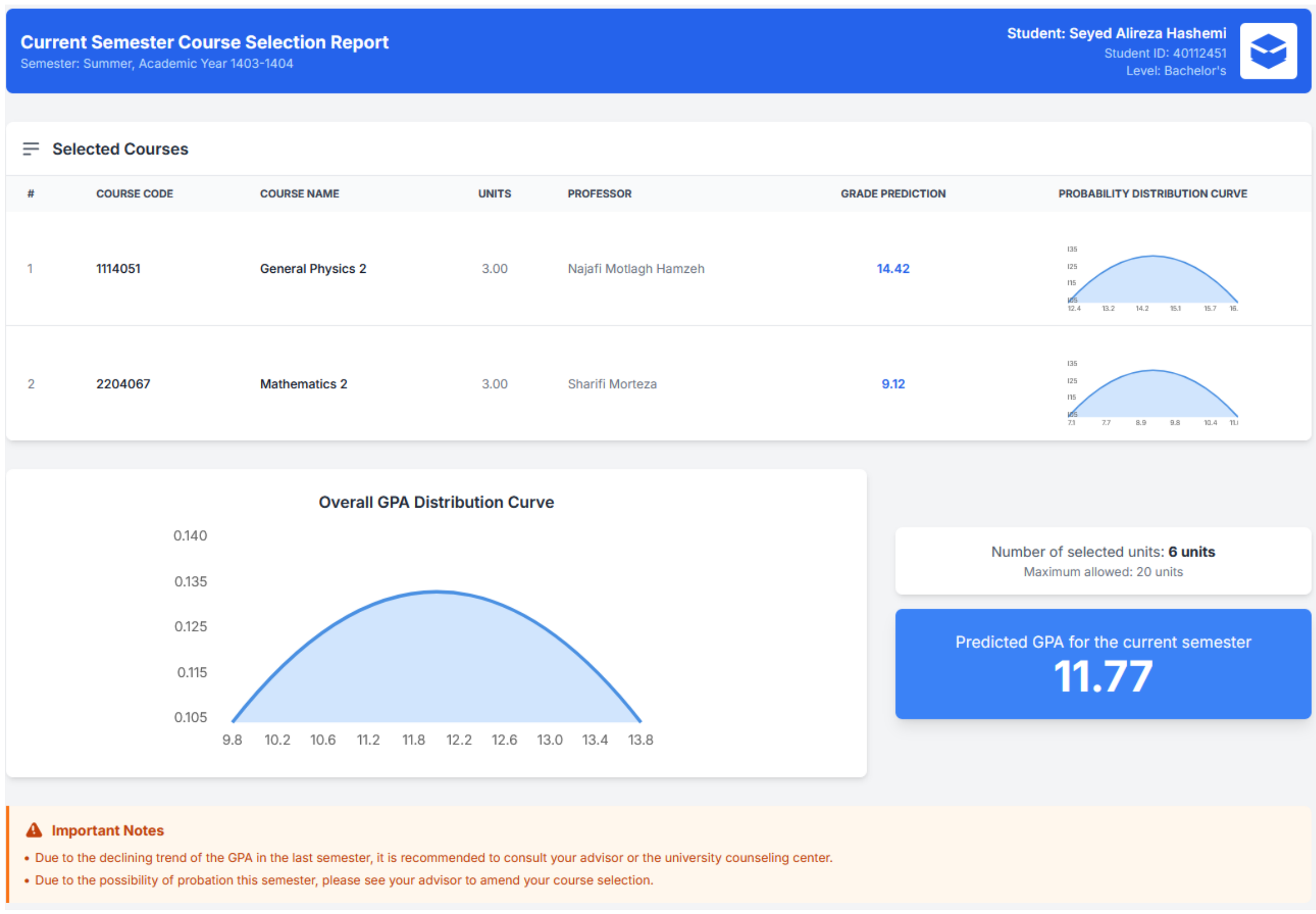

**figure 8. A sample of the user interface of the implemented tool in the Birjand University integrated system, showing predicted grades for selected courses, probable semester GPA, and rule-based alerts**

In addition to predicting grades for selected courses and calculating the semester GPA, the system also provides students a list of recommended courses. In this process, the system predicts the potential grades for other courses and the professors teaching them, the system provides a more detailed analysis. Based on these predictions, courses are classified into three difficulty levels (hard, moderate, and easy) allowing students to identify courses with the highest probability of achieving desirable grades as suitable options. Also, courses that may potentially lower the semester GPA are displayed in a different color. In this way, the student can choose from these options with more awareness to adopt a more optimal strategy to increase their semester GPA (Figure 9).

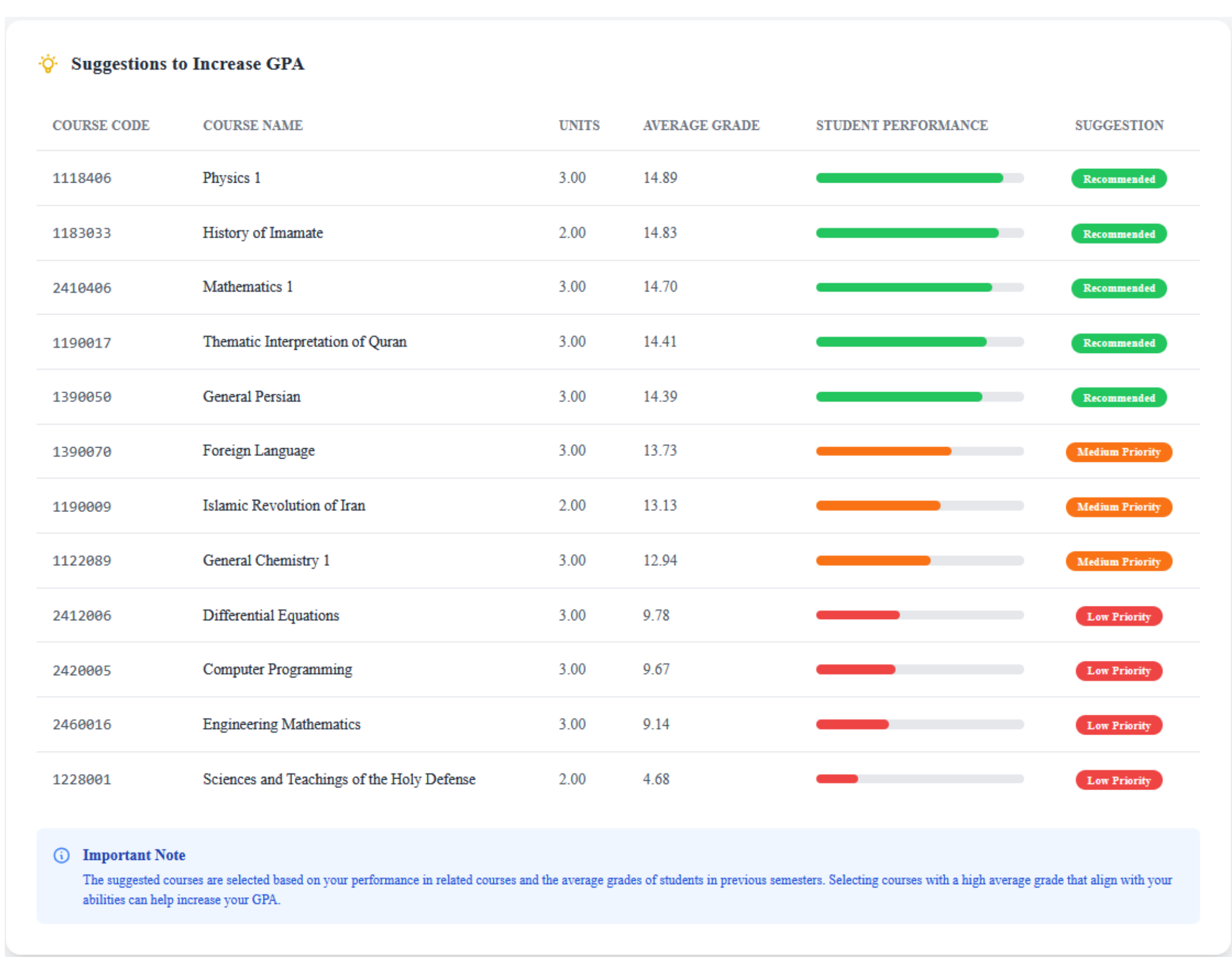

Suggestions to Increase GPA

| COURSE CODE | COURSE NAME | UNITS | AVERAGE GRADE | STUDENT PERFORMANCE | SUGGESTION |
|---|---|---|---|---|---|
| 1118406 | Physics 1 | 3.00 | 14.89 | | Recommended |
| 1183033 | History of Imamate | 2.00 | 14.83 | | Recommended |
| 2410406 | Mathematics 1 | 3.00 | 14.70 | | Recommended |
| 1190017 | Thematic Interpretation of Quran | 3.00 | 14.41 | | Recommended |
| 1390050 | General Persian | 3.00 | 14.39 | | Recommended |
| 1390070 | Foreign Language | 3.00 | 13.73 | | Medium Priority |
| 1190009 | Islamic Revolution of Iran | 2.00 | 13.13 | | Medium Priority |
| 1122089 | General Chemistry 1 | 3.00 | 12.94 | | Medium Priority |
| 2412006 | Differential Equations | 3.00 | 9.78 | | Low Priority |
| 2420005 | Computer Programming | 3.00 | 9.67 | | Low Priority |
| 2460016 | Engineering Mathematics | 3.00 | 9.14 | | Low Priority |
| 1228001 | Sciences and Teachings of the Holy Defense | 2.00 | 4.68 | | Low Priority |

Important Note
The suggested courses are selected based on your performance in related courses and the average grades of students in previous semesters. Selecting courses with a high average grade that align with your abilities can help increase your GPA.

**figure 9. User interface of the 'Course Recommendation for GPA Improvement' section, which prioritizes courses based on predicted grades**

In another component of this toolkit, a module based on a rule-based system has been designed and implemented to provide personalized recommendations and warnings for each student. This module generates actionable advice by analyzing the student's past academic data, predicted grades, progression trends, and the expert-defined rules embedded within the system. For example, the system can propose various strategies to help students improve their GPA or remind them of key points necessary for achieving higher GPA. Figure 10 shows an example of one of the suggestions provided by the system.

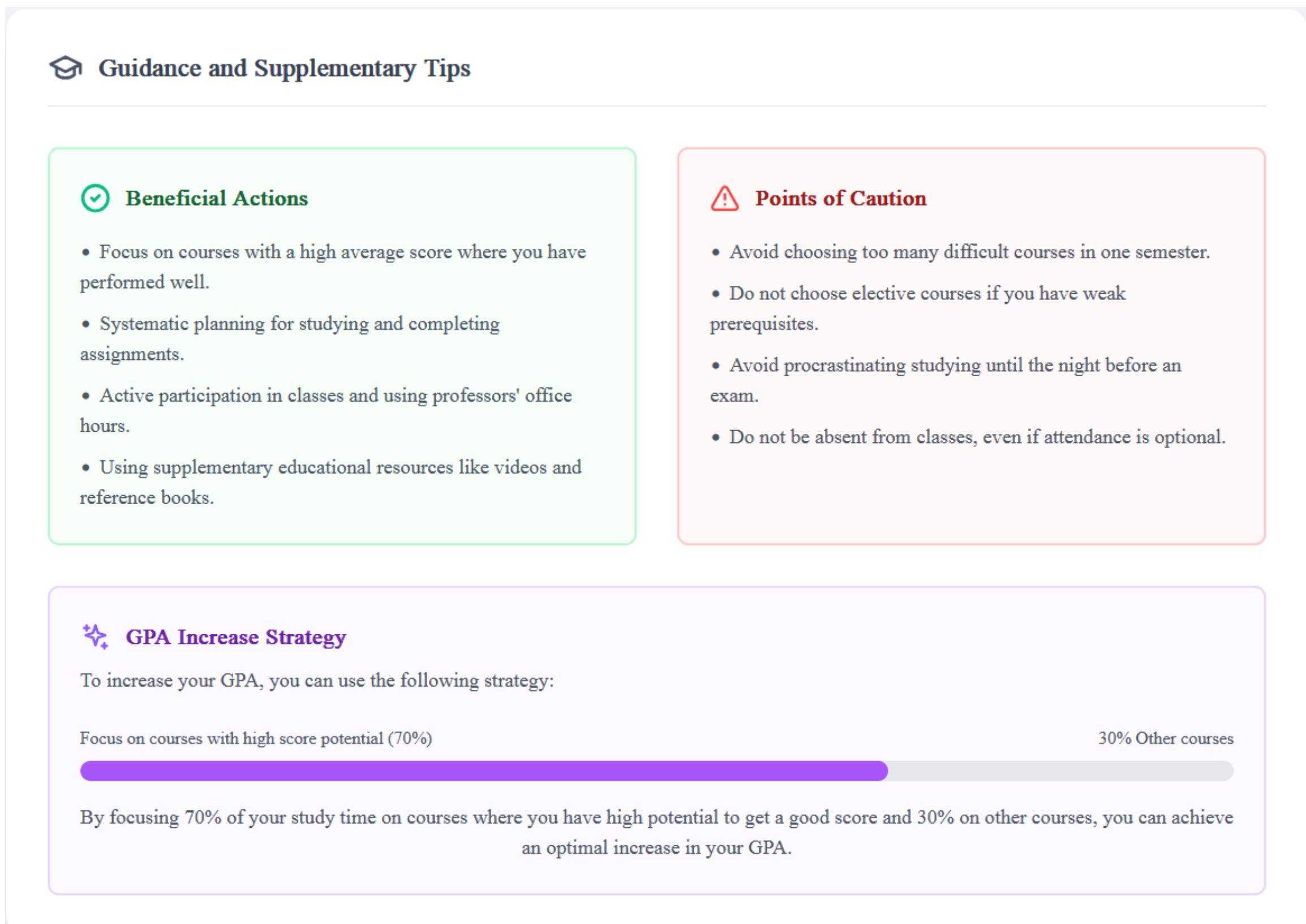

**Figure 10. The user interface of the "Recommender" module, featuring actionable guidance, cautionary alerts, and strategies for GPA improvement.**

### 4.5. Usability Evaluation

To assess the practical utility and user acceptance of the proposed framework, a usability survey was conducted with a focus group of 50 students who utilized the recommendation module during the course registration period. The survey was designed based on the Technology Acceptance Model (TAM), focusing on two key constructs: Perceived Usefulness (PU) and Perceived Ease of Use (PEOU).

The questionnaire consisted of five distinct questions, and participants were asked to rate their responses using a 5-point Likert scale ranging from 1 (Strongly Disagree) to 5 (Strongly Agree). The questions designed for this study are detailed in Table 8.

The results of the survey, detailing the number of students who selected each option for every question, are presented in Table 9.

**Table 8. Usability Survey Questions based on TAM**

| ID | Construct | Question Item |
|---|---|---|
| Q1 | Perceived Usefulness | The "GPA Prediction" feature helped me make more realistic course selections. |
| Q2 | Perceived Usefulness | The "Probation Risk Alert" and warnings provided by the system were effective and timely. |
| Q3 | Perceived Ease of Use | The explanations for course recommendations were clear, transparent, and easy to understand. |
| Q4 | Perceived Ease of Use | I found the system easy to navigate and use alongside the university portal. |
| Q5 | Behavioral Intention | I intend to use this intelligent assistant again in future semesters. |

**Table 9. Distribution of Student Responses (N=50)**

| Question | Strongly Disagree (1) | Disagree (2) | Neutral (3) | Agree (4) | Strongly Agree (5) | Positive Rate (4 & 5) |
|---|---|---|---|---|---|---|
| Q1 | 1 | 2 | 6 | 18 | 23 | 82% |
| Q2 | 0 | 1 | 5 | 15 | 29 | 88% |
| Q3 | 2 | 3 | 7 | 20 | 18 | 76% |
| Q4 | 0 | 2 | 4 | 14 | 30 | 88% |
| Q5 | 1 | 1 | 3 | 12 | 33 | 90% |

As demonstrated in Table 9, the overall feedback from the students was highly positive. Regarding Perceived Usefulness, 82% of participants (Agree and Strongly Agree) indicated that the GPA prediction feature significantly aided their decision-making process (Q1). Notably, the probation risk alerts (Q2) received the highest specific usefulness rating, with 88% of students affirming its effectiveness; this suggests that the proactive warning system is a critical component for at-risk students.

In terms of Perceived Ease of Use, 76% of respondents found the rule-based explanations clear and trustworthy (Q3). Although this is a positive result, the slight increase in "Neutral" and "Disagree" responses compared to other questions suggests that while the "White Box" approach is superior to black-box models, the textual explanations for complex rules could be further simplified in future versions. However, the general navigation (Q4) was rated very highly (88%), indicating successful integration with the existing university portal.

Finally, the Behavioral Intention (Q5) score was exceptional, with 90% of students expressing a desire to use the tool in future semesters. These quantitative results confirm that the proposed hybrid framework is not only technically robust but also widely accepted by the target user base, successfully bridging the gap between complex algorithmic predictions and practical, user-friendly academic advising.

## 5. Discussion

This study addresses the growing challenges of scalability, personalization, and transparency in academic advising by proposing and deploying a hybrid, multi-purpose intelligent framework. Unlike conventional approaches that focus exclusively on either grade prediction or rule-based advising, the proposed system integrates data-driven ensemble learning with a transparent rule-based expert system, resulting in a comprehensive and practically deployable solution for higher education institutions. The discussion of the findings is structured around three key dimensions: predictive performance, the synergy between artificial intelligence and expert knowledge, and practical implications for academic advising.

**5.1. Predictive Performance Analysis**

The experimental results demonstrate that the proposed cluster-based Stacking Ensemble model consistently outperforms individual base learners across all clusters and evaluation metrics. By partitioning the student population into homogeneous subgroups using Gaussian Mixture Models (GMM), the framework effectively reduces data heterogeneity and enables the training of specialized predictors tailored to distinct academic profiles. This design choice plays a critical role in improving predictive stability and generalization.

The aggregated performance of the ensemble model achieved an RMSE of 2.35 and an $R^2$ value of 0.53 on the unseen test set. While these values may appear moderate compared to controlled or small-scale datasets, they are highly meaningful in the context of real-world academic data. Student grading systems are inherently noisy and subjective, influenced by instructor-specific grading policies, course difficulty variations, and institutional constraints. Under such conditions, achieving an $R^2$ above 0.5 on a dataset exceeding 416,000 real academic records represents a strong and practically relevant outcome.

An important observation from the results is the comparatively lower performance of the MLP model across most clusters, including instances of negative $R^2$. This behavior aligns with existing findings in educational data mining, where tree-based ensemble methods often outperform neural networks on heterogeneous tabular datasets. In contrast to Random Forest and Gradient Boosting, MLP models are more sensitive to feature scaling, noise, and sparse categorical encodings, even when normalization is applied. Nevertheless, incorporating MLP into the ensemble contributes to model diversity, which ultimately enhances the robustness of the averaged stacking framework.

The statistical significance analysis further confirms the effectiveness of the proposed approach. The paired samples T-test, conducted under verified normality assumptions, yielded a p-value below 0.001, demonstrating that the reduction in prediction error achieved by the ensemble model is statistically significant rather than a result of random variation.

**5.2 Synergy of Artificial Intelligence and Expert Knowledge**

A central limitation of many machine-learning-based academic advising systems lies in their lack of interpretability and weak alignment with institutional regulations. Black-box models, despite their predictive power, often fail to gain trust among academic advisors and administrative staff. The proposed framework explicitly addresses this issue by integrating a rule-based expert system as a complementary decision layer.

In this hybrid architecture, the ensemble prediction model estimates a student's expected academic performance, while the rule-based engine acts as a pedagogical guardrail that enforces university regulations, prerequisite constraints, probation policies, and academic progression rules. This design ensures that system recommendations are not only statistically plausible but also policy compliant and pedagogically valid. By decoupling prediction from decision logic, the system achieves both flexibility and transparency two critical requirements for real-world adoption in higher education environments.

Moreover, the rule-based explanations provide students with clear and actionable feedback, reducing cognitive ambiguity and increasing confidence in the system's guidance. This explainability advantage distinguishes the proposed framework from purely data-driven or LLM-based advising systems, which often lack explicit control mechanisms and institutional accountability.

### 5.3. Practical Implications for Academic Advising

The successful implementation of this system within the University of Birjand portal offers practical implications that extend beyond algorithmic accuracy. This system shifts the academic advising paradigm across three key dimensions:

A) Scalability and Accessibility Traditional advising is constrained by office hours and faculty availability (synchronous constraints), often facing bottlenecks during course registration periods. The proposed system provides 24/7 asynchronous availability, significantly reducing the cognitive load on academic advisors. This allows faculty members to dedicate their limited time to complex psychological and career guidance issues, while the intelligent system manages routine regulatory monitoring and alerts.

B) Data-Driven Personalization Whereas human advisors often rely on intuition or limited memory of a student's history, this system analyzes 416,558 educational records to provide granular personalization. For instance, the system recommends courses where the student not only meets the prerequisites but also—based on the performance of similar peers in their specific cluster has the highest probability of success.

C) Shift from Reactive to Proactive Approaches Perhaps the most significant practical contribution is the transition from "Reactive Advising" (intervening after a student faces probation) to "Proactive Advising". By detecting a "Declining Trend" and predicting probation risk before course registration is finalized, the system issues timely warnings. This Instant Feedback empowers students to adjust their academic plans proactively, preventing academic failure—a preventative measure often missing in traditional systems.

### 5.4. Limitations

Despite the promising results, this study has limitations. The current dataset is specific to a single university, and evaluating its generalizability to other institutions requires further investigation. Additionally, while initial student feedback has been positive, a longitudinal study is recommended to quantitatively measure the long-term impact of the system on improving graduation rates and overall GPAs.

Another limitation of this study addresses the 'Cold Start' problem. Since the proposed hybrid model relies heavily on historical academic features (e.g., previous semester GPA and passed units) to predict performance and generate recommendations, first-semester students were excluded from the dataset due to the absence of prior university records. Consequently, the current system is optimized for students with at least one semester of academic history.

## 5.Conclusion

This study introduces and implements a hybrid and multi-functional framework designed for predicting course grades and providing intelligent course selection recommendations within a university setting. The proposed approach is based on three main stages: (1) Clustering student patterns using a Gaussian Mixture Model (GMM); (2) Training a specialized set of base models (Random Forest, Gradient Boosting, MLP) within each cluster; and (3) Aggregating the outputs via averaging and stacking methods, designed to reduce errors and enhance generalization capability. In addition to the prediction component, a Rule-Based engine is implemented in the system to ensure policy compliance and provide transparent explanations to the end-user, thereby increasing the system's explainability and operational acceptance.

The main contributions of this study can be summarized in three areas: (1) the design and evaluation of a cluster-based framework that, by separating different learning patterns, enables the training of specialized models capable of predicting students' grades; (2) the development and deployment of a rule-based system to provide academic recommendations to students within a web-based platform; and (3) the creation of a robust system for recommending courses for the upcoming semester to students, using both machine learning models and the rules defined within the proposed system.

Future work will aim to resolve the cold start problem for freshmen by incorporating pre-admission data, such as high school GPA and university entrance exam scores, into the predictive framework.

### Statements and Declarations

**Data availability statements:** The data used in this study are part of a university-owned database and are not publicly available due to institutional restrictions.

**Funding -** No funding was received for this research.

**Conflict of Interest**: All authors certify that they have no affiliations with or involvement in any organization or entity with any financial interest (such as honoraria; educational grants; participation in speakers' bureaus; membership, employment, consultancies, stock ownership, or other equity interest; and expert testimony or patent-licensing arrangements), or non-financial interest (such as personal or professional relationships, affiliations, knowledge or beliefs) in the subject matter or materials discussed in this manuscript.

### Generative AI use deceleration

The authors confirm that no artificial intelligence (AI) tools were used in the conceptualization, design, data collection, analysis, or interpretation of this study. Limited AI-based software assistance was employed solely for grammar and language editing purposes but was fully humanized.